%% file: sample-sigconf-authordraft.tex
\documentclass[sigconf,pbalance]{acmart}

\usepackage{lineno}
\usepackage{etoolbox}

\usepackage{latexsym}

\usepackage[T1]{fontenc}

\usepackage[utf8]{inputenc}

\usepackage{microtype}

\usepackage{graphicx}

\usepackage{listings} %
\usepackage{graphicx} %
\usepackage{fancyvrb} %
\usepackage{booktabs} 
\usepackage{amsmath}
\usepackage{multirow}
\usepackage{xspace}
\usepackage{subcaption}
\usepackage{enumitem} 
\usepackage{bbding}
\usepackage{bm}
\newcommand{\paratitle}[1]{\vspace{0.2cm}\noindent\textbf{#1}}
\newcommand{\ie}{\emph{i.e.,}\xspace}

\newcommand{\eg}{\emph{e.g.,}\xspace}

\newcommand{\ours}{\textbf{AMD}}

\AtBeginDocument{%
  }

\copyrightyear{2025}
\acmYear{2025}
\setcopyright{acmlicensed}\acmConference[CIKM '25]{Proceedings of the 34th ACM International Conference on Information and Knowledge Management}{November 10--14, 2025}{Seoul, Republic of Korea}
\acmBooktitle{Proceedings of the 34th ACM International Conference on Information and Knowledge Management (CIKM '25), November 10--14, 2025, Seoul, Republic of Korea}
\acmDOI{10.1145/3746252.xxxxxxx}
\acmISBN{xxx-x-xxxx-xxxx-x/2025/11}

\begin{document}

\title{Arrows of Math Reasoning Data Synthesis for Large Language Models: Diversity, Complexity and Correctness}

\author{Sirui Chen}
\authornote{Contribution during internship at Ant Group.}
\authornote{Equal contribution. \; \footnotemark[3]Corresponding author.}
\orcid{0009-0006-5652-7970}
\affiliation{
    \institution{Zhejiang University} 
    \city{Hangzhou} 
    \country{China}}
\email{chenthree@zju.edu.cn}

\author{Changxin Tian\footnotemark[2]}
\orcid{0000-0002-3013-9439}
\affiliation{
    \institution{Ant Group} 
    \city{Hangzhou} 
    \country{China}}
\email{tianchangxin.tcx@antgroup.com}

\author{Binbin Hu}
\orcid{0000-0002-2505-1619}
\affiliation{
    \institution{Ant Group} 
    \city{Hangzhou} 
    \country{China}}
\email{bin.hbb@antfin.com}

\author{Kunlong Chen}
\orcid{0009-0001-0961-1479} %
\affiliation{
    \institution{Ant Group} 
    \city{Hangzhou} 
    \country{China}}
\email{kunlong.ckl@antgroup.com}

\author{Ziqi Liu}
\orcid{0000-0002-4112-3504}
\affiliation{
    \institution{Ant Group} 
    \city{Hangzhou} 
    \country{China}}
\email{ziqiliu@antgroup.com}

\author{Zhiqiang Zhang\footnotemark[3]}
\orcid{0000-0002-2321-7259}
\affiliation{
    \institution{Ant Group} 
    \city{Hangzhou} 
    \country{China}}
\email{lingyao.zzq@antgroup.com}

\author{Jun Zhou}
\orcid{0000-0001-6033-6102}
\affiliation{
    \institution{Ant Group} 
    \city{Hangzhou} 
    \country{China}}
\email{jun.zhoujun@antgroup.com}

\renewcommand{\authors}{Sirui Chen, Changxin Tian, Binbin Hu, Kunlong Chen, Ziqi Liu, Zhiqiang Zhang, and Jun Zhou.}
\renewcommand{\shortauthors}{Chen and Tian, et al.}

\begin{abstract}
Enhancing the mathematical reasoning of large language models (LLMs) demands high-quality training data, yet conventional methods face critical challenges in scalability, cost, and data reliability. To address these limitations, we propose a novel program-assisted synthesis framework that systematically generates a high-quality mathematical corpus with guaranteed diversity, complexity, and correctness. This framework integrates mathematical knowledge systems and domain-specific tools to create executable programs. These programs are then translated into natural language problem-solution pairs and vetted by a bilateral validation mechanism that verifies solution correctness against program outputs and ensures program-problem consistency. We have generated 12.3 million such problem-solving triples. Experiments demonstrate that models fine-tuned on our data significantly improve their inference capabilities, achieving state-of-the-art performance on several benchmark datasets and showcasing the effectiveness of our synthesis approach.
\end{abstract}



\maketitle

\input{./sec1-intro}

\input{./sec2-method}

\input{./sec3-exp}

\input{./sec4-related}

\input{./sec5-con}

\bibliographystyle{ACM-Reference-Format}
\bibliography{custom.bib}

\appendix
\input{./sec8-genAI}

\end{document}

%% file: sec1-intro.tex
\section{Introduction}
Mathematical reasoning, a critical capability of large language models (LLMs) for real-world applications, has garnered substantial attention. Recent studies demonstrate that training LLMs on mathematical data enhances this capacity  \cite{azerbayev2023llemma,shao2024deepseekmath,team2025every}. 
However, with conventional mathematical corpora (e.g., web pages and textbooks) becoming progressively depleted, current approaches increasingly rely on either manual annotation \cite{wang2025mv} or more advanced LLMs to acquire consumable training data \cite{yu2023metamath,gou2023tora,wang2023mathcoder,luo2023wizardmath}. 
Despite their effectiveness, these approaches face significant challenges in cost and quality, posing non-negligible risks to LLM development:
(1) Manual annotation necessitates high-caliber annotators and incurs substantial costs when handling complex problem sets; 
(2) Existing LLM-synthesized corpora frequently exhibit unreliable correctness, with our random sampling revealing a 40\% error rate in existing synthetic datasets. 
Though such data may temporarily boost model performance, these quality issues undermine the long-term, sustainable enhancement of this capability. 

In this work, we aim to construct a scalable data synthesis pipeline to generate high-quality mathematical data with ensured diversity, complexity, and correctness for training LLMs. Our core approach leverages a systematic mathematical knowledge framework and mathematical tools to guarantee data quality through bilateral verification mechanisms. 
Unlike prior approaches that integrate tools or use program-of-thought paradigms, our framework leverages mathematical tools not to enhance models' tool-invocation capabilities, but to systematically generate natural language corpora for improving the mathematical reasoning abilities.

However, synthesizing high-quality mathematical corpora poses multifaceted challenges:
(1) \textbf{Diversity and Systematic Coverage}: Traditional methods relying on web corpora or benchmark-driven synthesis often overfit existing data distributions, failing to ensure both systematic coverage and sufficient diversity.
(2) \textbf{Complexity-Correctness Tradeoff}: While generating complex mathematical problems is crucial, 
increased problem complexity inherently amplifies unreliability in generated outputs and verification challenges.

To address these dilemmas, we propose a novel program-assisted generation approach 
that leverages executable programs to produce high-quality mathematical data.
For systematicity and diversity, we first construct a three-tier mathematical knowledge system (``education stage, subject, topic'') and associate each knowledge topic with mathematical tools. Through the combinatorial integration of knowledge topics and their corresponding tools, we generate executable programs that comprehensively cover the knowledge system. These programs are then combined with those derived from seed corpora to form a systematic and diverse collection of math programs. 
To balance complexity and correctness, we employ four mutation strategies (constraint, variable, constant, and code variations) on a mathematical program set, enhancing complexity and quantity. The mutated programs are translated back into natural language questions with corresponding solutions. 
Next, a bilateral validation mechanism ensures synthesis quality: LLM-generated solutions are verified against program execution outputs, effectively eliminating errors from the generation pipeline and ensuring program-problem consistency.
Using our approach, \ours, we synthesized over 12 million high-quality mathematical data samples. Experimental results demonstrate that fine-tuning LLMs with our data significantly enhances their reasoning abilities, often surpassing state-of-the-art methods on various evaluation datasets.

Our contributions are summarized as follows:
(1) We explore a novel mathematical data synthesis paradigm that leverages external tools to ensure the complexity and correctness of the synthetic corpus.
(2) We construct a comprehensive mathematical knowledge system and a corresponding mathematical toolkit to enable systematic synthesis of math reasoning data.
(3) We develop a scalable synthesis approach that simultaneously considers diversity, complexity, and correctness, generating over 12 million high-quality mathematical reasoning samples.
(4) Extensive experiments demonstrate both the effectiveness of our synthetic data and the superiority of our synthesis approach over conventional approaches.

%% file: sec2-method.tex
\begin{figure*}[ht!]
    \centering
    \includegraphics[width=0.85\linewidth]{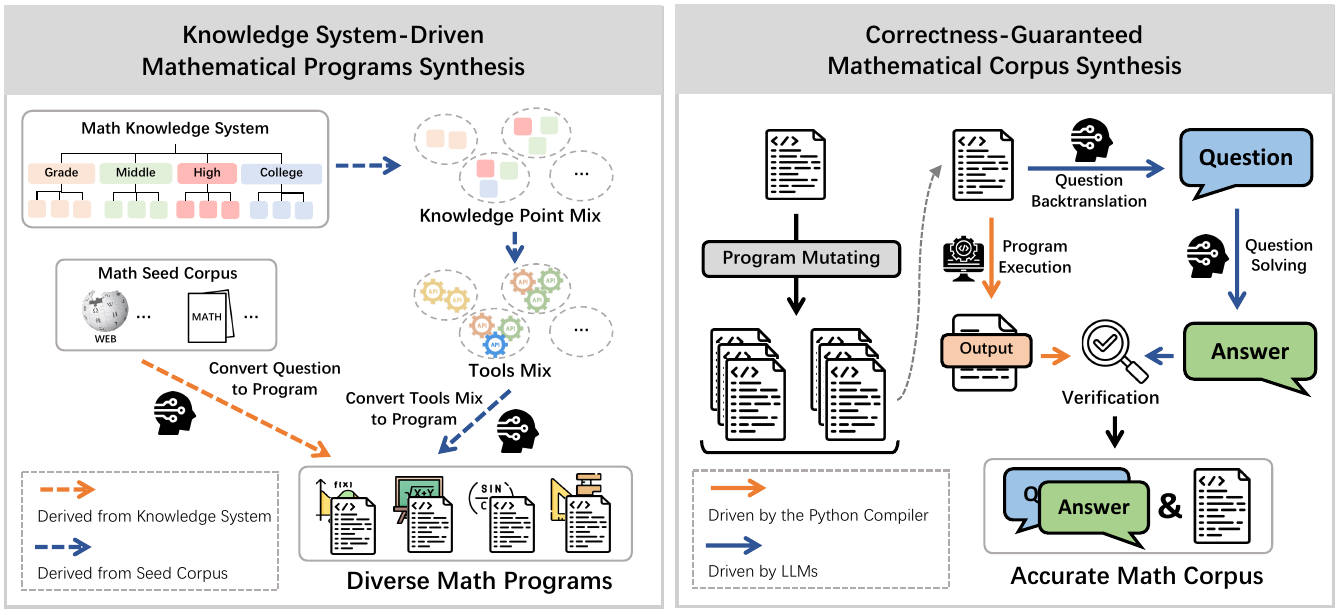}
    \caption{The pipeline of our approach. 
    We first develop a mathematical knowledge system mapping concepts to tools, then synthesize diverse math programs by integrating these tools with a seed corpus.
    Next, these programs are mutated for increased complexity and translated into natural language questions.
    Finally, an LLM generates solutions using Chain-of-Thought (CoT) reasoning, which are then verified against program execution results for correctness.
    }
    \label{fig:main}
\end{figure*}

\section{Approach}

Different from tool-integrated LLM approaches \cite{gou2023tora,yue2023mammoth}, we focus on synthesizing a correctness-guaranteed math corpus to enhance the LLM's intrinsic mathematical reasoning, rather than its function-calling abilities. 
As shown in Fig.~\ref{fig:main}, we first construct a knowledge system that maps mathematical concepts to tools. These tools are integrated into executable programs, forming a diverse program set along with those derived from a seed mathematics corpus. We then mutate these programs to generate more complex variations and translate them into natural language questions. 
Finally, we solve these questions using LLMs and conduct bilateral verification between the LLM-generated solutions and the program execution results to filter out a correctness-guaranteed corpus.

Formally, our goal can be summarized as: \textit{Based on a comprehensive mathematical knowledge system $\mathcal{K}$ and seed questions $\mathcal{S}$, we construct a diverse and complex set of math programs $\mathcal{C}=\{c\}$. From this, we create a correctness-guaranteed math corpus $\mathcal{D} = \{(c, q, a)\}$, where $c$ is a synthetic math program, $q$ and $a$ are the corresponding math question and solution, respectively.}

\subsection{Knowledge System-driven Mathematical Programs Synthesis. }
To construct the knowledge system, we integrate two foundational resources: K-12 mathematics textbooks
and the mathematical taxonomy from the Chinese Library Classification System. 
We then exploit the integration capabilities of GPT-4 as well as the knowledge of human experts to develop a comprehensive knowledge system.
As a result, our knowledge system $\mathcal{K}$ organizes more than 250 key topics in mathematics into a three-tier hierarchy of education stage, subject and topic (such as \textit{College → Linear Algebra → Eigenvalues}), ensuring curricular alignment and academic rigor.

\paratitle{Constructing Systematic Mathematical Toolkit.}
Building upon the knowledge system $ \mathcal{K} $, we create an automated pipeline to map mathematical topics to tools. 
For each topic $  k \in \mathcal{K}  $, we select 10-50 representative problems from curricular materials and open-source math corpora. GPT-4 generates programs to solve these problems using scientific computing libraries (\eg \texttt{SymPy}), ensuring standard API usage and step-by-step derivations. We then extract the related APIs, $\{t\}_k$, from these programs through syntax pattern mining, establishing a mapping from topic $k$ to tools $\{t\}_k$.
Finally, we aggregate the related APIs from all knowledge topics to obtain the tool set  $\mathcal{T} = \bigcup_{k \in \mathcal{K}} \{t\}_k$, where $ \mathcal{T} $ consists of over 100 APIs. 
These APIs form a Mathematical Toolkit, organized by Topic (e.g., Matrix Diagonalization) and Atomic Operation (e.g., \texttt{numpy.linalg.qr}),  
bridging the knowledge system with computational practice.

\paratitle{Synthesizing Programs via Tool Combinations. }
Inspired by ``Sentence Building Games'', we combine tools from different math topics in $\mathcal{K}$ to construct executable programs.
Specifically, each generation process begins by sampling 1-3 mathematical topics from our structured knowledge system $\mathcal{K}$ through stratified random sampling, forming a topic combination $(k_1, k_2, ...)$, which maintains curriculum coherence through constraint-based selection. 
Each topic $k_i$ is then mapped to its corresponding code tools through our toolkit, generating a tool combination $({t}_{k_1}, {t}_{k_2}, ...)$. 
Finally, the program generator \texttt{LLM}$_P$ uses the topic and tool combinations, along with a prompt, to generate a program as follows:
$$\mathcal{C}_K = \bigcup_{(k_1, ...) \sim \mathcal{K}} \texttt{LLM}_P((k_1, ...), ({t}_{k_1}, ...), \pi_{p})$$
where $\mathcal{C}_K$ 
are programs generated from our knowledge system and $\pi_p$ is the program generation prompt.
Additionally, to enhance real-world relevance, we sample about 100 thousand problems from standard benchmarks to form a seed question set $\mathcal{S}$. These problems are solved using programs to generate an additional program set:
$$\mathcal{C}_S = \bigcup_{q^\prime \in \hat{\mathcal{S}}} \texttt{LLM}_P(q^\prime, \pi_{p\prime}),$$
where $\pi_{p\prime}$ denotes the program solution generation prompt.%
By merging the above program sets, the final program set $\mathcal{C} = \mathcal{C}_K \cup \mathcal{C}_S$ is a diverse collection that fulfills two key objectives: 1) systematic coverage through knowledge graph traversal, 2) empirical grounding via real-world problem distribution matching.

\subsection{Program-driven Mathematical Corpus Synthesis. }
Instead of augmenting data directly, we first enhance program complexity through mutation and then translate these programs into new question-solution pairs. The correctness of these pairs is then ensured through bilateral validation, comparing program execution results against the LLM-generated solutions.

\paratitle{Mutating Programs for Greater Complexity. }
Our synthesized program set $\mathcal{C}$ combines two generation sources (the knowledge system and seed questions) to ensure both systematic coverage and diversity. However, empirical result reveals inherent limitations in direct LLM-generated programs $\mathcal{C} = \{c\}$, which often suffer from oversimplification and insufficient quantity.
To overcome these limitations, we develop a mutation strategy with four specialized operators:
1) \textbf{Constraint Mutation}: deepens problem complexity by adding/modifying constraints (\eg ``$0<x$'' → ``$0 < x < 5$''). 
2) \textbf{Variable Mutation}: introduces new variables or substitutes existing ones (\eg linear → multivariate equations). 
3) \textbf{Constant Mutation}: perturbs numerical constants or reformat expressions (\eg $±10\%$ to $±300\%$).  
4) \textbf{Code Mutation}: inserts control structures or replaces API calls. 
Formally, our mutation model $\texttt{LLM}_M$ processes each program $c \in \mathcal{C}$ through $k$ iterative refinements:
$$
\mathcal{C}_M = \bigcup_{c\in\mathcal{C}} \bigcup_{m=1}^k \texttt{LLM}_M(c, \pi_{m})
$$
where $\pi_m$ denotes mutation templates guiding specific complexity enhancements.
Finally, we combine the original and mutated programs to form the final program set $\widehat{\mathcal{C}} = \mathcal{C} \cup \mathcal{C}_M$, which not only ensures the diversity of the programs but also significantly enhances their complexity and quantity.

\paratitle{Guaranteeing Correctness via Bilateral Verification. }
The programs in $\mathcal{C}$ can be translated back into question-solution pairs, constructing a mathematical corpus in natural language. 
However, this strategy raises challenges from two aspects: 
1) Consistency: the reverse-engineered natural language problem should align with the code. 
2) Validity: the problem should be clearly defined and solvable. 
Due to the current limitations of LLMs, these issues cannot be fully resolved. 
Moreover, our mutation strategy for mathematical programs exacerbates these challenges, leading to unreliable synthetic data. 
To address this, we adopt a bilateral verification mechanism to ensure the correctness of the synthesized data, \ie execute the programs and compare the outputs with the LLM-generated solutions. 
Our verification mechanism operates through three core steps for each program $c \in \mathcal{C}$:
(1) \textbf{Question Generation}: produce question $q_c$ via LLMs, \ie $q_c = \texttt{LLM}_G(c, \pi_q)$. 
(2) \textbf{Answer Extraction}: derive answers through dual channels: 
\begin{align*}
    s_c &= \texttt{LLM}_G(c, \pi_s) \\
    a_c &= \texttt{regex\_extract}(s_c) \\
    o_c &= \texttt{Interpreter}(c).
\end{align*}
In practice, we sample multiple solutions for each question to scale the quantity of data.
(3) \textbf{Cross-Verification}: construct verified corpus: $\mathcal{D} = \{ (c, q_c, s_c) \ | \ c \in \widehat{\mathcal{C}} \land a_c \equiv o_c \}$. 
Note that errors at any step from ``$ c \rightarrow q_c \rightarrow s_c \rightarrow a_c $'' can lead to inconsistencies between $a_c$ and $o_c$. 
Thus, our two-sided verification mechanism not only ensures the correctness of semantic translation (code $\rightarrow$ question) and logical derivation (question $\rightarrow$ solution). This process ultimately results in a correctness-guaranteed math corpus $\mathcal{D}$.

\subsection{Fine-tuning using Synthetic Data}
Our pipeline produces 12.3 million program-question-solution triples, with 2.6 million derived from the knowledge system and 9.7 million from seed corpus expansion. 
Among these samples, there are 1.8 million unique program-question pairs, averaging about 6.8 solutions per pair. 
Following established methodologies, we filter out instances with 10-gram overlaps in both inputs and outputs from the test sets of downstream evaluation tasks. The filtered synthesis data are then used to fine-tune open-source LLMs, enabling them to predict solutions based on the given problems.

%% file: sec3-exp.tex
\setlength{\floatsep}{5pt plus 2pt minus 2pt}
\setlength{\textfloatsep}{5pt plus 2pt minus 2pt}
\setlength{\intextsep}{5pt plus 2pt minus 2pt}

\section{Experiments}

\subsection{Experimental Settings}

We follow existing work \cite{neuro-symbolic}, applying LoRA to perform supervised fine-tuning on three pre-trained models: LLaMA3-8B \cite{grattafiori2024llama}, Mistral-7B \cite{jiang2023mistral7b}, and Deepseek-Math-7B \cite{shao2024deepseekmath}. To evaluate the effectiveness of AMD, we assess the accuracy before and after fine-tuning on four benchmarks: GSM8K \cite{GSM8K}, MATH \cite{Math},  Minerva\_Math \cite{Minerva}, and SVAMP \cite{SVAMP}. We use existing mathematical data synthesis methods, MathGenie \cite{mathgeine}, as comparative methods. To ensure a fair evaluation of data quality, we randomly sample 50,000 instances from the publicly available training datasets of each method for fine-tuning.

\subsection{Overall Performance}

\begin{table}[t]
\scriptsize
\centering
\caption{Performance comparison among our method and other mathematical data synthesis methods.}
\label{tab: Performance Comparison}
\vspace{-0.3cm}
\begin{tabular}{ccccccc}
\toprule
Base & Models & GSM8K & MATH & Minerva & SVAMP \\
\midrule
\multirow{3}{*}{LLaMA3-8B} & - & 55.5 & 17.3 & 18.2 & 69.1 \\
 & MathGenie & 23.7 & 19.1 & 17.8 & 30.5 \\
 & AMD & \textbf{58.5} & \textbf{23.5} & \textbf{19.2} & \textbf{76.1} \\
\midrule
\multirow{3}{*}{Mistral-7B} & - & 39.8 & 11.8 & 11.4 & 63.9 \\
 & MathGenie & \textbf{45.8} & 14.5 & 13.6 & 71.3 \\
 & AMD & 45.7 & \textbf{16.6} & \textbf{18.2} & \textbf{71.6} \\
\midrule
\multirow{3}{*}{Deepseek-Math-7b} & / & 63.7 & 32.3 & 29.4 & 74.0 \\
 & MathGenie & \textbf{66.9} & 34.9 & 30.6 & \textbf{82.8} \\
 & AMD & 66.6 & \textbf{37.7} & \textbf{32.6} & 80.8 \\
 \bottomrule
\end{tabular}%
\end{table}

Table \ref{tab: Performance Comparison} presents a comprehensive performance comparison between AMD and the baseline mathematical data synthesis approach across three base models and four mathematical reasoning benchmarks. The results demonstrate that AMD consistently outperforms MathGenie in most settings, particularly showing significant gains on weaker base models like LLaMA3-8B. While MathGenie occasionally achieves marginal advantages in specific configurations, AMD exhibits more robust performance across different model architectures and task domains. The consistent superior performance across heterogeneous evaluation metrics confirms AMD's effectiveness in generating high-quality mathematical training data.

\subsection{Ablation Study}

\begin{table}[t]
\centering
\scriptsize
\caption{Ablation study of our data derived from the knowledge system/seed corpus.}
\label{tab: ablation-derived from}
\vspace{-0.3cm}
\begin{tabular}{cccccccc}
\toprule
Base & \begin{tabular}[c]{@{}c@{}}Knowledge\\ System\end{tabular} & \begin{tabular}[c]{@{}c@{}}Seed\\ Corpus\end{tabular} & GSM8K & MATH & Minerva & SVAMP \\
\midrule
\multirow{3}{*}{LLaMA3-8B} & $\checkmark$ & $\checkmark$ & 58.5 & 23.5 & 19.2 & 76.1 \\
 & $\checkmark$ &  & 56.1 & 23.1 & 19.2 & 76.5 \\
 &  & $\checkmark$ & 57.8 & 23.1 & 19.0 & 77.8 \\
\midrule
\multirow{3}{*}{Mistral-7B} & $\checkmark$ & $\checkmark$ & 45.7 & 16.6 & 18.2 & 71.6 \\
 & $\checkmark$ &  & 45.3 & 16.2 & 17.8 & 70.1 \\
 &  & $\checkmark$ & 46.1 & 16.8 & 17.8 & 69.1 \\
\midrule
\multirow{3}{*}{Deepseek-Math-7B} & $\checkmark$ & $\checkmark$ & 66.6 & 37.7 & 32.6 & 80.8 \\
 & $\checkmark$ &  & 66.3 & 38.3 & 32.6 & 79.9 \\
 &  & $\checkmark$ & 68.1 & 38.1 & 35.4 & 78.6\\
 \bottomrule
\end{tabular}%
\end{table}

\begin{table}[t]
\scriptsize
\centering
\caption{Ablation study of our data with different languages.}
\label{tab: ablation-language}
\vspace{-0.3cm}
\begin{tabular}{ccccccc}
\toprule
Base & EN & CN & GSM8K & MATH & Minerva & SVAMP \\
\midrule
\multirow{3}{*}{LLaMA3-8B} & $\checkmark$ & $\checkmark$ & 58.5 & 23.5 & 19.2 & 76.1 \\
 & $\checkmark$ &  & 56.1 & 23.2 & 17.6 & 75.5 \\
 &  & $\checkmark$ & 57.8 & 22.0 & 20.2 & 76.5 \\
\midrule
\multirow{3}{*}{Mistral-7B} & $\checkmark$ & $\checkmark$ & 45.7 & 16.6 & 18.2 & 71.6 \\
 & $\checkmark$ &  & 45.3 & 15.4 & 17.8 & 69.2 \\
 &  & $\checkmark$ & 45.5 & 16.7 & 19.6 & 69.6 \\
\midrule
\multirow{3}{*}{Deepseek-Math-7B} & $\checkmark$ & $\checkmark$ & 66.6 & 37.7 & 32.6 & 80.8 \\
 & $\checkmark$ &  & 66.3 & 37.8 & 34.0 & 79.8 \\
 &  & $\checkmark$ & 67.7 & 38.2 & 33.4 & 79.7 \\
 \bottomrule
\end{tabular}%
\end{table}

The ablation studies of two data synthesis pathways and two languages are shown in Table \ref{tab: ablation-derived from} and Table \ref{tab: ablation-language}. When combining knowledge system-derived data and seed corpus expansion, all models achieve optimal or near-optimal performance across benchmarks. The ablation study of different languages shows that our method is effective in both Chinese and English, and combining data from both languages will not reduce the performance of the model.

%% file: sec4-related.tex
\section{Related Work}

Large language models (LLMs) have demonstrated remarkable capabilities in many fields \cite{achiam2023gpt,zhao2023survey,team2025every}, but their performance in mathematical reasoning remains inconsistent \cite{lu2022survey, ahn2024large}.
To improve their performance on mathematical reasoning, current methods often employ chain of thoughts (CoT) hints on general models \cite{CoT}, or enhance models by pre-training or fine-tuning with specialized mathematical datasets to improve their performance \cite{azerbayev2023llemma,shao2024deepseekmath}.
However, the large amount of mathematical data required by these methods is difficult to collect manually, so there has been a surge in methods for automatically synthesizing high-quality mathematical datasets in recent years.
Jiuzhang \cite{jiuzhang30} creates a knowledge distillation dataset to train a small LLM for generating mathematical content.
MathGenie \cite{mathgeine} generates mathematical problems with corresponding codes to verify the correctness of the solutions, but it lacks diversity because its data is not combined and developed from a mathematical perspective.
Neuro-Symbolic Data Generation \cite{neuro-symbolic} introduces a symbolic system to convert and mutate the original text into new symbolic problems, enhancing the diversity of the problems with correctness by verifying the solutions with their symbolic representation. However, its mutation methods are confined to some pre-defined patterns, limiting the complexity of the generated content.

%% file: sec5-con.tex
\section{Conclusion}

In this work, we propose a novel mathematical data synthesis paradigm, build a comprehensive mathematical knowledge system and a corresponding mathematical toolkit. We implement knowledge system-driven mathematical program synthesis and correctness-guaranteed mathematical corpus synthesis, replacing traditional corpus synthesis that completely relies on LLMs with external tool generation that considers diversity, complexity, and correctness. Experiments have demonstrated the effectiveness of our synthetic data and that our synthesis method is superior to traditional methods.

%% file: sec8-genAI.tex
\section{GenAI Usage Disclosure}

In accordance with ACM's guidelines on generative AI usage disclosure, we provide a comprehensive account of how generative AI tools were utilized in our paper.

\paratitle{{GenAI Usage in Research Methodology}}
We utilized GPT-4 as part of our research methodology, specifically for:

\begin{itemize}
    \item \textbf{Knowledge System-driven Mathematical Programs Synthesis:}
    GPT-4 is used to integrate information from K-12 mathematics textbooks and the Chinese Library Classification System, working alongside human experts to develop our comprehensive mathematical knowledge system covering over 250 key topics.
    GPT-4 also serves as the program generator  ($\texttt{LLM}_P$) to create executable mathematical programs by combining tools from different mathematical topics.
    
    \item \textbf{Program-driven Mathematical Corpus Synthesis:}
    GPT-4 functions as the mutation model ($\texttt{LLM}_M$) to enhance program complexity through our four specialized mutation operators, and is employed as the generation model ($\texttt{LLM}_G$) to translate programs into natural language questions and generate corresponding solutions.
\end{itemize}

\paratitle{{GenAI Usage in Experimental Evaluation}}
GPT-4 is employed as an evaluation tool in our experiments to assess the accuracy of base models and fine-tuned LLMs across different datasets. Specifically, we use GPT-4 to compare LLM-generated solutions against ground truth answers, determining correctness through semantic equivalence rather than exact string matching. This automated evaluation approach ensures consistent and scalable assessment across our large-scale experiments.